\documentclass{article}

    \usepackage[preprint]{neurips_2023}

\usepackage[utf8]{inputenc} %
\usepackage[T1]{fontenc}    %
\usepackage{hyperref}       %
\usepackage{url}            %
\usepackage{booktabs}       %
\usepackage{amsfonts}       %
\usepackage{nicefrac}       %
\usepackage{microtype}      %
\usepackage{xcolor}         %
\usepackage{color-edits}
\usepackage{algorithm}
\usepackage{algorithmicx}
\usepackage{algpseudocode} %

\usepackage{bm}
\usepackage{amsmath}
\usepackage{xcolor}
\usepackage{hyperref}
\hypersetup{
    colorlinks=true,
    linkcolor=blue,
    filecolor=magenta,      
    urlcolor=cyan,
    pdftitle={Overleaf Example},
    pdfpagemode=FullScreen,
    }

\usepackage{amssymb}
\usepackage{graphicx}
\usepackage{xspace}

\usepackage{times}
\usepackage{latexsym}
\usepackage{graphicx}
\usepackage{float}
\usepackage{amsmath}
\usepackage[T1]{fontenc}
\usepackage{multirow}
\usepackage[utf8]{inputenc}
\usepackage{xspace}

\usepackage{booktabs}

\usepackage{hyperref}
\usepackage{url}

\usepackage{braket}
\usepackage{tikz}
\usepackage{pgfplots}
\pgfplotsset{compat=1.18}
\usepackage{amssymb}
\usepackage{xcolor}
\usepackage{svg}
\usepackage{tikzscale}
\usepackage{graphicx}
\usepackage{longtable}
\usepackage{xcolor}
\usepackage{wrapfig}
\usepackage{nameref}
\usepackage{lscape, rotating}
\usepackage{enumitem}
\usepackage{listings}
\usepackage{afterpage}

\usepackage{ulem}

\usepackage{wrapfig}
\usepackage{framed}
\usepackage{subcaption}

\usepackage{booktabs}
\usepackage{tabularx}

\addauthor{kb}{purple}

\title{A Surprising Failure? \\ Multimodal LLMs and the NLVR Challenge}

\author{%
  Anne Wu, Kianté Brantley, and Yoav Artzi\\
  Department of Computer Science and Cornell Tech, Cornell University\\
  \texttt{\{annewu, yoav\}@cs.cornell.edu}, \texttt{kdb82@cornell.edu} \\
}

\begin{document}

\maketitle

\begin{abstract}

This study evaluates three state-of-the-art MLLMs --- GPT-4V, Gemini Pro, and the open-source model IDEFICS --- on the compositional natural language vision reasoning task NLVR. 
Given a human-written sentence paired with a synthetic image, this task requires the model to determine the truth value of the sentence with respect to the image. 
Despite the strong performance demonstrated by these models, we observe they perform poorly on NLVR, which was constructed to require compositional and spatial reasoning, and to be robust for semantic and systematic biases.

\end{abstract}

\section{Introduction}\label{sec:intro}

Recent multimodal large language models (MLLMs) are demonstrating strong performance across a range of vision and language tasks~\citep{gpt4v2023, team2023gemini, yang2023dawn}. 
In the past, vision and language models have repeatedly demonstrated strong results, only to be qualified later because of relying on semantic biases rather than robust processing of the visual input and its correspondence to the input text~\citep[inter alia]{agrawal2017c, cirik2018visual, kojima2020learned}.
This question remains open for MLLMs. 
We use the Natural Language Visual Reasoning~\citep[NLVR;][]{suhr2017corpus} task to study this question.
NLVR uses simple geometrical shapes, and was generated to be robust to systematic and semantic biases. 
We examine three recent MLLMs: GPT-4V, Gemini Pro, and IDEFICS. 
Largely, we observe poor performance on NLVR across all three, showing that the fundamental spatial and compositional challenges that NLVR presents remain unaddressed even by contemporary MLLMs.

\section{Task Background: NLVR}\label{sec:task}

\begin{figure}[h]
\captionsetup[subfigure]{justification=centering}
    \centering
    \begin{subfigure}[c]{0.45\textwidth}
        \includegraphics[width=\linewidth]{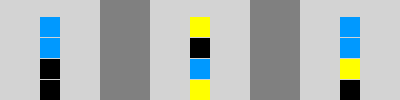}
        \caption*{\textit{There are two towers with the same height but their base
is not the same in color}}
        \label{fig:nlvr_fig:tower}
    \end{subfigure}\hfill
    \begin{subfigure}[c]{0.45\textwidth}
        \includegraphics[width=\linewidth]{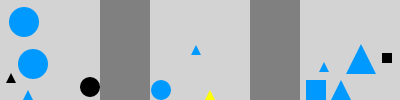}
        \caption*{\textit{There is a box with items of only 2 different colors and a small black triangle touching the wall}}
        \label{fig:nlvr_fig:scatter}
    \end{subfigure}
        \caption{Examples of sentence-image pairs from the NLVR corpus. The left sentence has a truth value of \texttt{True} with respect to the \texttt{Tower} image. The right sentence has a truth value of \texttt{False} with respect to the \texttt{Scatter} image.}
    \label{fig:nlvr_fig}
\end{figure}

The NLVR task is to determine whether a sentence is \texttt{True} or \texttt{False} with regards to a given image. 
There are two types of images: \texttt{Tower}, where the images only contain squares stacked in towers with up to 4 squares, and \texttt{Scatter}, which contains scattered objects of different sizes, with 1--8 objects in each box. \autoref{fig:nlvr_fig} shows two examples of sentence-image pairs from the corpus.\footnote{\autoref{table:stats} in the Appendix provides data statistics.} 

The goal of NLVR is to assess the ability of models to demonstrate fundamental compositional and spatial reasoning skills. 
It uses simple shapes with little semantic meaning, for example using simple geometric shapes rather than everyday objects. 
NLVR was generated and annotated to be robust to systematic biases. 
The annotation process focused the sentence writing task on precise understanding of spatial relations and their composition. It is often the case in NLVR that relatively small details are crucial to correctly complete the task (e.g., ``small black triangle'' in the right image of \autoref{fig:nlvr_fig}), and that reasoning about objects or image regions in isolation in insufficient. 
\cite{suhr2017corpus} describes the corpus creation process.

\section{Experimental Setup}\label{sec:setup}

We evaluate three models on the \texttt{Test-P} split, which includes  5{,}940 examples.

\subsection{Models}\label{sec:setup:models}

\paragraph{GPT-4 Turbo with Vision (GPT-4V)} GPT-4 Turbo is the second generation of GPT-4~\citep{gpt4v2023}. GPT-4V  is referred to as \texttt{gpt-4-vision-preview} through the API. %

\paragraph{Gemini Pro}

Gemini are a family of multimodal models~\citep{team2023gemini} trained on text, image, audio and video. We evaluated using the second-largest model of the first version, as the Ultra model was not yet publicly available.

\paragraph{IDEFICS} IDEFICS~\citep{laurenccon2023obelisc} is an open-source vision-language model following the Flamingo~\citep{alayrac2022flamingo} design, based on publicly available data and models. 
We use the 9B parameters instructed version. We evaluate using both the frozen model, and a version fine-tuned on the NLVR training split.

\subsection{Model Evaluation Details}\label{sec:setup:model_eval}

We queried GPT-4V and Gemini Pro via the LiteLLM interface~\citep{litellm}, using the OpenAI API for GPT-4V and the Google Vertex AI API for Gemini Pro. The GPT-4V queries are done between Dec. 29, 2023 -- Jan. 11, 2024, and the Gemini Pro queries are done between Dec. 29, 2023 -- Jan. 2, 2024.
When prompting, we generate with greedy search using a temperature of 0.

For IDEFICS fine-tuning, we quantize the 9B instruct model to 4-bits precision using QLoRA~\citep{dettmers2023qlora}.
We use a learning rate of 3e-6 with a batch size of 32.

\subsubsection{Prompts Selection}\label{sec:setup:prompt_select}

We experiment with both zero- and five-shot prompting for each model.

For zero-shot prompting, to get the best possible performance out of the models, we randomly sample a subset of 100 examples from the Test-P split, and experiment with a set of manually-designed 14 candidate prompts for both GPT-4V and Gemini Pro, and eight candidate prompts for IDEFICS.  
We then select the best performing prompt for each model based on accuracy. 
The candidate prompts are built with different prompt engineering methods (e.g., delimiters, chain-of-thought, etc.).

For five-shot prompting, we randomly sample a subset of 20 examples and ran six candidate prompts, starting from the best zero-shot prompt for each model. 
We selected the prompt that had the highest accuracy on this small set to test on the complete test data, separately for each model. 
We ensure both labels are represented in the five examples. 
To select the examples, for each \texttt{Test-P} example, we randomly sample 5 examples from the training set. 
For \texttt{Tower} test examples, we choose \texttt{Tower} training examples, and the same for \texttt{Scatter}. 

The selected prompts are different for each of model, and are provided in \autoref{sec:prompts}.

\section{Results \& Analysis}\label{sec:exps:results}

\autoref{table:accuracies} shows the overall accuracy on \texttt{Test-P}. Performance is overall relatively low, significantly below human performance, and below existing state-of-the-art~\citep{zheng2021km4}, which used a relatively complex custom architecture. 

Among the prompting approaches, GPT-4V with zero-shot prompting achieves the best overall performance, similar to the accuracy of the fine-tuned IDEFICS. The best performing prompt selected for this result is a detailed prompt with step-by-step instructions and chain-of-thought.

Few-shot prompting only improves the performance for Gemini Pro, for which the zero-shot performance was low. The performance degrades for the other two models. One possibility is that the examples are randomly sampled from the training split and contain different sentences, which may not be directly beneficial for the specific test example. 
For GPT-4V, the best prompt uses chain-of-thought, but as we don't have the intermediate reasoning annotation for the training examples, only the label is provided.

Gemini Pro shows bias towards predicting \texttt{False} for both the zero- and few-shot settings, but it improves in the few-shot case. The predictions of IDEFICS are biased towards \texttt{True} in the zero-shot setting, but changed to \texttt{False} in the few-shot setting. 

\begin{table*}[t]
\centering
\footnotesize
\setlength{\tabcolsep}{5pt}
\begin{tabular}{lcccccc}
\toprule
Model & Technique & Overall Accuracy & TP & FN & FP & TN \\
\midrule
Human & - & 95{.}4 & - & - & - & -\\
Majority class \citep{suhr2017corpus} & - & 56{.}2 & - & - & - & -\\
\cite{zheng2021km4} & - & 78{.}3 & - & - & - & -\\
\midrule
GPT-4V & Zero-shot & 59{.}9 & 2195 & 1141 & 1239 & 1365\\
Gemini Pro & Zero-shot & 49{.}9 & 659 & 2677 & 300 & 2304\\
IDEFICS & Zero-shot & 55{.}9 & 3271 & 65 & 2555 & 49\\
GPT-4V & Five-shot & 58{.}0 & 2248 & 1088 & 1406 & 1198\\
Gemini Pro & Five-shot & 51{.}5 & 1363 & 1973 & 907 & 1697\\
IDEFICS & Five-shot & 45{.}1 & 738 & 2598 & 664 & 1940\\
IDEFICS & Fine-tuned & 59{.}7 & 2144 & 1200 & 1192 & 1404\\
\bottomrule
\end{tabular}
\caption{Accuracy on the \texttt{Test-P} split. TP is true positive, FN is false negative, etc.}
\label{table:accuracies}
\vspace{-10pt}
\end{table*}

\begin{figure}[h]
    \centering
    \includegraphics[width=10cm]{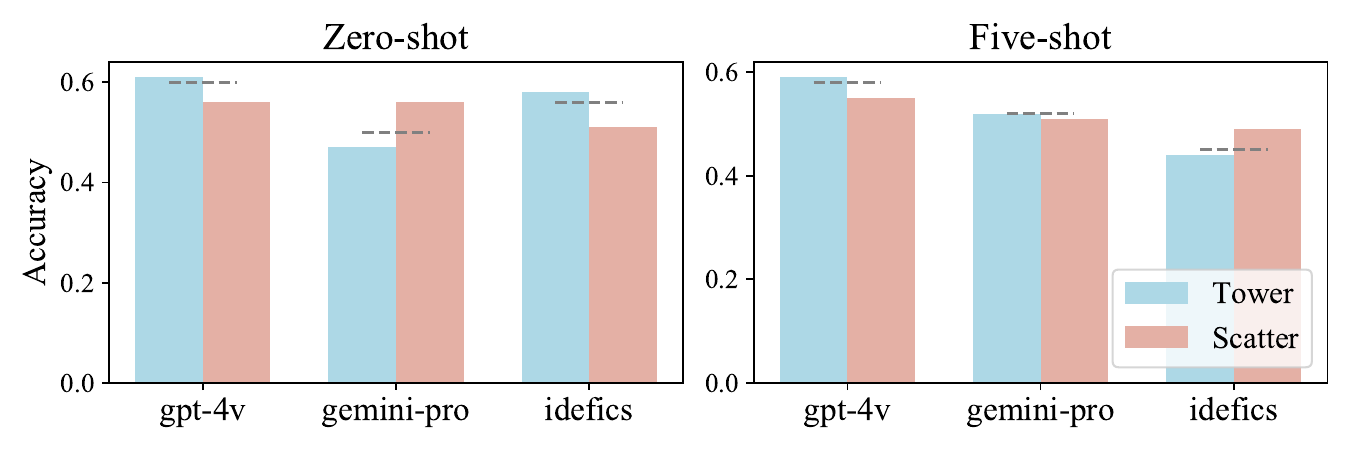}
    \caption{\texttt{Test-P} accuracies with zero- and five-shot  prompting, split by image type (Tower or Scatter).}
    \label{fig:accuracy-by-tower-scatter}
\end{figure}

\autoref{fig:accuracy-by-tower-scatter} shows the accuracy for both zero- and five-shot prompting, by image type. In the zero-shot setting, both GPT-4V and IDEFICS perform better on Tower, which is simpler and has a more structured visual representation. In the few-shot setting, Gemini Pro does better on Tower.

\section{Conclusion}\label{sec:conclusion}

We compared GPT-4V, Gemini Pro, and IDEFICS on the compositional natural language visual reasoning task NLVR. 
We found that with only prompting, either model remains far from the human accuracy. Fine-tuning the open-source IDEFICS model improved the performance, but there remains a large room for improvement.

This study has several limitations. 
Results can change and hamper reproducibility as APIs are constantly being updated and/or deprecated. 
Although we select the best prompt among a set of candidate prompts, the use of different prompts can influence the results, and they could significantly change with further prompt engineering techniques. 
Finally, we do not have access to the training data of GPT-4V and Gemini Pro, the results shown only indicate how the models perform on this task. We do not have the ability to conduct analysis such as data leakage.

\section*{Acknowledgement}

This research was supported by ARO W911NF21-1-0106, NSF under grant No. 1750499, and a gift from Open Philanthropy.

\bibliographystyle{plainnat}
\bibliography{custom}

\appendix
\section{Data Statistics}

\autoref{table:stats} shows data statistics for the NLVR corpus. 
A sample in the NLVR corpus consists of a sentence and an image. The boxes in each image can be permuted, and there are in total six sentence-image pairs for each sample in each split. The statistics include those permutations.
We evaluated on \texttt{Test-P}. For the few-shot approach, we draw examples from the \texttt{Train} split. For the fine-tuning approach, we fine-tune on \texttt{Train}.

\begin{table*}[ht]
\centering
\setlength{\tabcolsep}{5pt}
\begin{tabular}{lcc}
\toprule
Split & Unique sentences & Examples \\
\midrule
Train & 3{,}163 &  74{,}460 \\ %
Dev & 267 & 5{,}934 \\
Test-P & 266 & 5{,}940 \\
Test-U & 266 & 5{,}910 \\
\midrule
Total & 3{,}962 & 92{,}244 \\
\bottomrule
\end{tabular}
\caption{Data statistics for NLVR.\protect\footnotemark}
\label{table:stats}
\vspace{-10pt}
\end{table*}

\footnotetext{The number of examples for Test-P is updated according to the actual number of examples in the split.}

\section{Selected Prompts}\label{sec:prompts}

\autoref{tab:zero-shot} shows example prompts used for zero-shot model evaluation.
\autoref{tab:five-shot} shows example prompts used for five-shot model evaluation.

\begin{table}[h]
\centering
\setlength{\tabcolsep}{5pt}
\begin{tabularx}{\textwidth}{lX}
\toprule
\textbf{Model} & \textbf{Prompt} \\
\midrule
GPT-4V & You will be provided with a sentence and an image.\newline\newline Follow these steps to answer if the sentence is true or false with respect to the image:\newline 1. Look at the sentence (given after the line starting with "Sentence: ").\newline 2. Look at the image.\newline 3. Check if the sentence is true or false with respect to the image. Enclose all your work for this step within triple quotes (""").\newline 4. Answer with the choice you think is correct. Possible answer choices: True / False.\newline\newline Think step by step and do careful reasoning. Don't decide until you have checked both the sentence and the image carefully.\newline\newline Sentence: [sentence]\newline Answer: \newline [image]\\
\midrule
Gemini Pro & Is the sentence [sentence] true or false with respect to the following image?\newline [image] \\
\midrule
IDEFICS & [image]\newline Is the sentence [sentence] true or false with respect to the image? Only answer with `False' or `True'\newline Answer: \\
\bottomrule
\end{tabularx}
\caption{Example of prompts used for zero-shot model evaluation}
\label{tab:zero-shot}
\end{table}

\begin{table}[h]
\centering
\setlength{\tabcolsep}{5pt}
\begin{tabularx}{\textwidth}{lX}
\toprule
\textbf{Model} & \textbf{Prompt} \\
\midrule
GPT-4V & You will be provided with a sentence and an image.\newline\newline Follow these steps to answer if the sentence is true or false with respect to the image:\newline 1. Look at the sentence (given after the line starting with "Sentence: ").\newline 2. Look at the image.\newline 3. Check if the sentence is true or false with respect to the image. Enclose all your work for this step within triple quotes (""").\newline 4. Answer with the choice you think is correct. Possible answer choices: True / False.\newline\newline Think step by step and do careful reasoning. Don't decide until you have checked both the sentence and the image carefully.\newline\newline Sentence: [sentence]\newline Answer: [label] \newline [image] \newline\newline Sentence: [sentence]\newline Answer: [label] \newline [image] \newline\newline ... \newline\newline Sentence: [sentence]\newline Answer: \newline [image]\\
\midrule
Gemini Pro & Is the sentence [sentence] true or false with respect to the following image?\newline [image]\newline Answer: [label] \newline\newline Is the sentence [sentence] true or false with respect to the following image?\newline [image]\newline Answer: [label] \newline\newline ... \newline\newline Is the sentence [sentence] true or false with respect to the following image?\newline [image] \\
\midrule
IDEFICS & [image]\newline Is the sentence [sentence] true or false with respect to the image? Only answer with `False' or `True'\newline Answer: [label] \newline\newline [image]\newline Is the sentence [sentence] true or false with respect to the image? Only answer with `False' or `True'\newline Answer: [label] \newline\newline ... \newline\newline [image]\newline Is the sentence [sentence] true or false with respect to the image? Only answer with `False' or `True'\newline Answer: \\
\bottomrule
\end{tabularx}
\caption{Example of prompts used for five-shot model evaluation}
\label{tab:five-shot}
\end{table}

\end{document}